\documentclass[preprint,12pt]{elsarticle}

\usepackage{url}
\usepackage{amssymb}
\usepackage{makecell}
\usepackage{subfig}
\usepackage{graphicx}
\usepackage{multirow}
\usepackage{float}
\usepackage{diagbox}[2011/11/22]
\usepackage{epsfig}
\usepackage{threeparttable}
\usepackage{color}
\usepackage{multirow}
\usepackage{bm}
\usepackage{soul}
\usepackage{algorithm}

\usepackage{algpseudocode}
\usepackage{amsmath}
\usepackage{amssymb}

\usepackage{multirow}

\usepackage{amssymb}

\usepackage{diagbox}
\journal{EAAI}

\begin{document}

\begin{frontmatter}



\title{Purifying Real Images with an Attention-guided Style Transfer Network for Gaze Estimation}


\author[label1]{Yuxiao Yan}
\ead{yuxiaoyan@dlmu.edu.cn}
\author[label1]{Yang Yan}
\ead{yanyang@dlmu.edu.cn}
\author[label1]{Jinjia Peng}
\ead{pengjinjia@dlmu.edu.cn}
\author[label1]{Huibing Wang}
\ead{huibing.wang@dlmu.edu.cn}
\author[label1]{Xianping Fu\corref{cor1}}
\ead{fxp@dlmu.edu.cn}

\cortext[cor1]{Xianping Fu is corresponding author.}

\address[label1]{Information Science and Technology College, Dalian Maritime University, \\ Dalian 116026, China}
\address{}

\begin{abstract}

Recently, the progress of learning-by-synthesis has proposed a training model for synthetic images, which can effectively reduce the cost of human and material resources. However, due to the different distribution of synthetic images compared to real images, the desired performance cannot be achieved. Real images consist of multiple forms of light orientation, while synthetic images consist of a uniform light orientation. These features are considered to be characteristic of outdoor and indoor scenes, respectively. To solve this problem, the previous method learned a model to improve the realism of the synthetic image. Different from the previous methods, this paper try to purify real image by extracting discriminative and robust features to convert outdoor real images to indoor synthetic images. In this paper,
we first introduce the segmentation masks to construct RGB-mask pairs as inputs, then we design a attention-guided style transfer network to learn style features separately from the attention and bkgd(background) region , learn content features from full and attention region. Moreover, we propose a novel region-level task-guided loss to restrain the features learnt from style and content. Experiments were performed using mixed studies (qualitative and quantitative) methods to demonstrate the possibility of purifying real images in complex directions. We evaluate the proposed method on three public datasets, including LPW, COCO and MPIIGaze. Extensive experimental results show that the proposed method is effective and achieves the state-of-the-art results.

\end{abstract}


\end{frontmatter}


\section{Introduction}\label{sec:introduction}

Appearance-based gaze estimation has recently progressed under outdoor conditions by using a large-scale real-image training data set with annotations through recent rises in high-capacity deep convolution networks. However, annotating training data sets requires a lot of manual labor. To solve this problem, a training model on a synthetic image is preferred because the annotations are automatically available. But this solution has a drawback, the distribution between the real image and the synthetic image is quite different. The distribution of synthetic images is more prone to indoor lighting, with slight variations depending on the synthesis method. On the other hand, due to the interference of light and other external factors, the distribution of real images is more complicated (prone to outdoor lighting), making the distribution of real images difficult to learn. Therefore, using synthetic images for training, the effects of testing in real scenes or on real image datasets will not be satisfactory. One solution is to attenuate the distribution of real images by improving the simulator, which can be expensive and time consuming. Another solution is to use unmarked actual data to improve the authenticity of the synthetic image from the simulator, such as SimGANs \cite{Ashish2017}, these methods only learn the global features without considering local features\cite{wu2018deep1}\cite{wu20193}. In the gaze estimation task, after realization with simGANs, the shape of the pupil or the edge of the pupil might by changed, the gaze estimation error will be increased due to the wrong pupil center location. Thus, these methods cannot be applied to outdoor (field) scenes due to its weak training time and adaptability to different situations in the field.

In a different manner, we try to purify real image by extracting discriminative and robust features to convert outdoor real images to indoor synthetic images. Synthetic images is more regular and easy to learn, meanwhile, the annotations are automatically available.

Similar to traditional style transfer network \cite{Gatys2015} \cite{lifeifei2016}, we need to capture the synthetic image style and transfer the real image with synthetic images' style but retaining the content of real image. Gatys et al.\cite{Gatys2015} proposed a method of using neural networks to capture artistic image styles and transfer them to real-world photos, Feifei Li et al.\cite{lifeifei2016} proposed using a perceptual loss function to train feed-forward networks for image transformation tasks. The main difference with Gatys et al.\cite{Gatys2015} and Feifei Li et al.\cite{lifeifei2016} is that images for gaze prediction need more-precise content information and more emphasis on image spatial arrangement of reservations.

To avoid changing the shape of the pupil or the edge of the pupil, we propose an attention-guided\cite{wu2018deep2}\cite{wu2018and} style transfer network to learn both local and global features. The way to handle local features is to obtain the attention region (pupil or iris) by segmentation. Fortunately, with the rapid development of deep learning based image segmentation methods including FCN\cite{Long2015}, SegNet\cite{Badrinarayanan2015}, U-net\cite{Ronneberger2015}, Mask R-CNN\cite{He2017}, we can obtain much better mask. With the mask, we can divide the eye image into two region: attention region (pupil, iris) and background region (skin and others). meanwhile, the attention region can contribute to gaze estimation in two respects. Firstly, the attention region can help removing the effects of texture features such as skin on pupil learning during feature learning in pixel-level. This can greatly improve the robustness of gaze estimator under various of background conditions. Secondly, the attention region contains pupil center information which can regarded as the most important features in gaze estimation.

To learn the style and content information from synthetic images, we First introduce the segmentation masks to construct RGB-mask pairs as inputs\cite{ACycle-Consistent}, then we design a attention-guided style transfer network to learn style features separately from the attention and bkgd(background) region , learn content features from full and attention region. For feature extraction, our work is most directly related to the work initiated by Gates et al.\cite{Gatys2015}. The feature map of the deep convolutional neural network with differentiated training is used to achieve the breakthrough performance of the transfer of painting style. We train a feed-forward feature extraction network for image transformation tasks. Our network aims to learn as much as possible on the premise of synthetic distribution, to minimize the loss of content transmission, and to solve the problem of insufficient spatial alignment information caused by the gram matrix. To achieve this goal, we propose a loss network with a novel task-guided loss, the attention region ,background region and full image region will be calculated in different task.

Our contributions are presented in this paper in four folds:

1. We took the first step to consider the attention region in style transfer task and propose an attention-guided style transfer network to purify the real image, making it similar to indoor conditions while retaining annotation information. Different with previous work in refining the synthetic images with global features, we purified the real images with local and global features.

2. We proposed a loss network with a novel task-guided loss function to maximize the content of the real image and the distribution of the synthetic image.

3.Our network not only considers the RGB color channel, but uses the segmentation masks to construct RGB-mask pairs as inputs. We learned style features separately from the attention and bkgd(background) region and learned content features from full and attention region.

4. We proposed a hybrid research method (qualitative and quantitative) for experiments on two tasks. The results show that the proposed architecture significantly purified the real image compared with the baseline methods. Meanwhile, We achieve the state-of-the-art results on gaze estimation task.

\begin{figure*}[!htbp]
\centering
 \begin{minipage}[]{1\textwidth}
    \centering
     \includegraphics[width = 1\textwidth,angle=0]{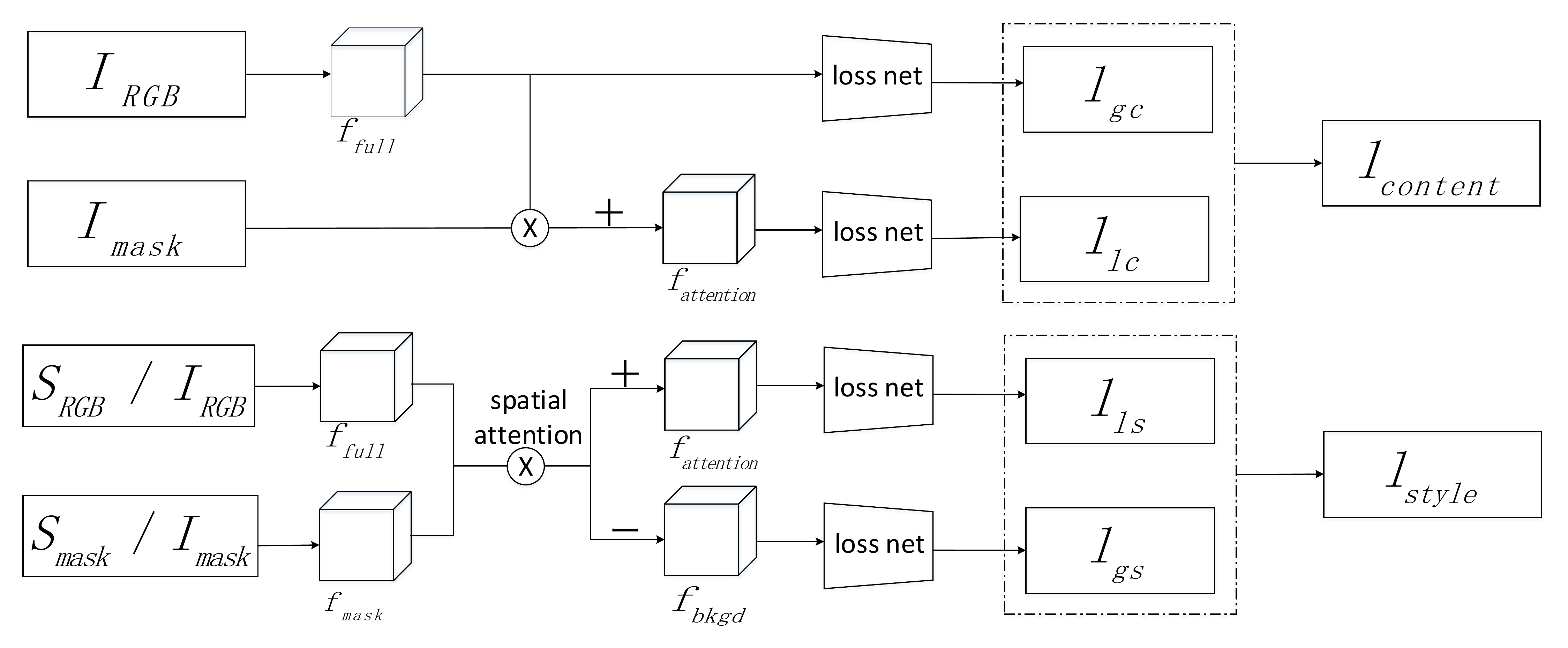}
 \end{minipage}
    \caption{Framework of proposed attention-guided style transfer network for gaze estimation. It contains three multi-scale stages and a loss net to learn final features. There are three main streams which extracted from different regions of image , i.e. , the full-stream $f_{full}$, the attention stream $f_{attetion}$, the background stream $f_{bkgd}$. In the lower part is the attention subnet which generates a pair of attention-region maps and background-region maps. A region-level task-guided loss is implemented on the features learnt from three streams for two task : retaining the content of input image $I_{RGB}$ and transferring the style from style image $S_{RGB}$ with the input image mask $I_{mask}$ and style image mask $S_{mask}$. }
    \label{figoverview}
\end{figure*}

\section{Related Works}\label{sec:relatedworks}

\subsection{Appearance-based gaze estimation}

The appearance-based approach is believed to work better under natural light. Recent studies aim to better represent the appearance, Lu et al. \cite{lu2014adaptive} proposed a low fifteen-dimensional feature extraction method to calculate the gray value and the percentage of each area. Wang et al.\cite{wang2016appearance} introduced a deep feature extracted from convolutional neural networks which has sparse characters and provides a effective solution for gaze estimation.

\subsection{Eye image synthesis}
There are four main categories of eye image synthesis methods: Optical Flow\cite{lu2015gaze}\cite{Wang2014hierarchical}, 3D eye reconstruction\cite{Sugano2014learning}\cite{Wood2015rendering}, Model-based method \cite{Wood2016a3d} and GANs (Generative Adversarial Networks)\cite{Shrivastava2016learning}. These methods tune parameters to obtain high resolution eye images, which are coincide with the ground truth situation. Shrivastava et al.\cite{Shrivastava2016learning} used GANs to generate synthetic eye images using unlabeled real data and learnt a refiner model that improves the realism of these synthetic images.

\subsection{Learning-by-synthesis}
Learning-based methods perform well in appearance-based gaze estimation but require large amounts of training data. Wang et al.\cite{Wang2014hierarchical} proposed an appearance-based gaze estimation method by supervised adaptive feature extraction\cite{AIterative}\cite{ALearning}\cite{AMultiview}\cite{ARobust}\cite{ASemantic}\cite{wang2012multimodel} and hierarchical mapping model\cite{ACycle-Consistent}\cite{ADeep}\cite{wang2018multiview} \cite{wang2017effective}\cite{wu2019cross}\cite{yang2017person}, during which appearance synthesis method is proposed to increase the sample density. Sugano et al.\cite{Sugano2014learning} presented a learning-by-synthesis approach for appearance-based gaze estimation and trained a 3D gaze estimator by a large amount of cross-subject training data.

\subsection{Style Transfer}

Previous methods learn a model to improve the realism of synthetic images, instead we take the first step to purify real images to weaken the influence of light and convert the distribution of outdoor real image to that of indoor synthetic image. This can be seen as a style transfer task, global style transfer algorithms process an image by applying a spatially-invariant transfer function. Reinhard et al.\cite{Reinhard2001} match the means and standard deviations between the input and reference style image after converting them into a decorrelated color space. Local style transfer algorithms based on spatial color mappings are more expressive and can handle a broad class of applications such as transfer of artistic edits\cite{Gatys2015}\cite{Selim2016}, weather and season change\cite{Gardner2015}. Similar to Gatys et al.\cite{Gatys2015}, which proposed a novel approach using neural networks to capture the style of artistic images and transfer it to real-world photographs, and Feifei Li et al.\cite{lifeifei2016} which proposed the use of perceptual loss function for training feed-forward networks for image transformation tasks. Our approach not only uses high-level feature representations of images from hidden layers of the VGG convolutional network to separate and reassemble content and style but also trains feed-forward networks to better calculate the loss of content and style.

\section{Proposed Method}\label{sec:Proposedmethod}

In brief, we reviewed the style transfer approach introduced by Gatys et al.\cite{Gatys2015} and Feifei Li et al. \cite{lifeifei2016} which transfer the style image $S_{RGB}$ into the input image $I_{RGB}$ by learning the global features. Gatys et al.\cite{Gatys2015} generated the stylized image $O_{RGB}$ by minimizing the objective function consisting of a content loss and a style loss.The Objective function can be represented as:

\begin{equation}
L_{total}=\sum_{l=1}^L\alpha_{l}L_{content}^l+\sum_{l=1}^L\beta_{l}L_{style}^l
\end{equation}
where L is the total number of convolutional layers and $l$ indicates the $l$-th convolutional layer of the deep convolutional neural network. $\alpha_{l}$ and $\beta_{l}$ are the weights to configure layer preferences. Each layer with $N_{l}$ distinct filters has $N_{l}$ feature maps each of size $M_{l}$, where $M_{l}$ is the height times the width of the feature map. So the responses in each layer l can be stored in a matrix $F[\cdot]\in R^{N_{l}\times M_{l}}$ where $F[\cdot]_{ij}$ is the activation of the $i^{th}$ filter at position $j$ in each layer $l$. The content loss, denoted as $L_{content}$, is simply the mean squared error between $F_{l}[O_{RGB}]\in R^{N_{l}\times M_{l}}$ and $F_{l}[I_{RGB}]\in R^{N_{l}\times M_{l}}$.

\begin{equation}
L_{content}^l=\frac{1}{N_{l}M_{l}}\sum_{ij}(F_{l}[O_{RGB}]-F_{l}[I_{RGB}])_{ij}^2
\end{equation}
The style loss, denoted as Lstyle, can be represented as:

\begin{equation}
L_{style}^l=\frac{1}{N_{l}^2}\sum_{ij}(G_{l}[O_{RGB}]-G_{l}[I_{RGB}])_{ij}^2
\end{equation}
Gram matrix $G_{l}[\cdot]$ is defined as the inner product between the vectored feature maps which is $F_{l}[\cdot]F_{l}[\cdot]^{T}\in R^{N_{l}\times N_{l}}$. Feifei Li et al.\cite{lifeifei2016} consists of two components: an image transformation network $f_{W}$ and a loss network $\phi$, $f_{W}$ is a deep residual convolutional network with weights $W$, it transforms $I_{RGB}$ into $O_{RGB}$ via mapping $O_{RGB} = f_{W}(I_{RGB})$. Loss network $\phi$ is used to minimize the loss between $O_{RGB}$ and $I_{RGB}$ , $O_{RGB}$ and $S_{RGB}$ with perceptual loss method. Loss between $O_{RGB}$ and $I_{RGB}$ is denoted as feature reconstruction $\ell_{feat}$ which can be represent as :

\begin{equation}
\ell_{feat}^{\phi,j} (O_{RGB},I_{RGB})=\frac{1}{C_{j}H_{j}W_{j}}\|\phi_{j}(O_{RGB})-\phi_{j}(I_{RGB})\|_{2}^2
\end{equation}
where j is a convolutional layer and  $\phi_{j}(\cdot)$ is a feature map of shape $C_{j}\times H_{j}\times W_{j}$. Loss between $O_{RGB}$ and $S_{RGB}$ is denoted as style reconstruction loss which is the squares Frobenius norm of the difference between the Gram matrices(similar with \cite{Gatys2015}) of $O_{RGB}$ and $I_{RGB}$:

\begin{equation}
\ell_{style}^{\phi,j}(O,S)=\|G_{j}^{\phi}(O)-G_{j}^{\phi}(S)\|_{2}^{2}
\end{equation}
 The Gram matrix can be computed efficiently by reshaping $\phi_{j}(\cdot)$ into a matrix $\psi$ of shape $C_{j}\times H_{j}\times W_{j}$; then $G_{j}^{\phi}(\cdot)=\frac{\psi\psi^{T}}{C_{j}\times H_{j}\times W_{j}}$. Image $O_{RGB}$ is generated by solving the problem

 \begin{equation}
 \begin{split}
O_{RGB}=\arg\min_{I_{RGB}}\alpha\ell_{feat}^{\phi,j}(O_{RGB},I_{RGB})\\+\beta\ell_{style}^{\phi,j}(O_{RGB},S_{RGB})+\theta\ell_{TV}(I_{RGB})
\end{split}
\end{equation}
where $\alpha$,$\beta$,$\theta$ are scalars.

As shown in Figure 1, there are three multi-scale stages and a loss net to learn final features. It contains three multi-scale stages and a loss net to learn final features. There are three main streams which extracted from different regions of image , i.e. , the full-stream $f_{full}$, the attention stream $f_{attetion}$, the background stream $f_{bkgd}$. The full-stream $f_{full}$ learns features from the raw images. Meanwhile, the attention stream $f_{attetion}$ and the background stream $f_{bkgd}$ are learned attention features and background features with attention maps. The attention maps are generated by the attention subnet. Although the features of these three streams are learnt from same input, they actually learnt features quite different features, for example, the one learnt from backgrounds which contains almost none useful information related to pupil center but contains the illumination information that can determine the difficulty of gaze estimation. To this end , a task-guided of constrains are added to restrain three features, with different task, the streams are designed to retaining the content of input image $I_{RGB}$ and transferring the style from style image $S_{RGB}$ with the input image mask $I_{mask}$ and style image mask $S_{mask}$.

For the given input(style) image and mask pair (RGB-Mask), the framework first produced the feature map $f_{full}$ and $f_{mask}$, then the attention subnet produces attention maps or a pair of contrastive attention maps and background maps as its source inputs. Through the loss net, we can separately calculate the content loss and the style loss. In the following subsections, we describe the details of the proposed method.

\subsection{Attention Subnet}

Given the input(style) image pair (RGB-Mask) as inputs, the attention subnet then produces attention maps which can be denoted as

\begin{equation}
att^{+}=\sigma(weight*(f_{full},f_{mask})+b)
\end{equation}
where $\sigma$ is the sigmoid function, weight and $b$ are the convolutional filter weights and bias. In the contrary, the background maps denoted as $att^{-}$, $att^{+}$ and $att^{-}$ constitute a contrastive attention pair, for each location $(i,j)$ which in the pair of attention maps and backgrounds maps should meet the constraint:

\begin{equation}
att^{+}(i,j)+att^{-}(i,j)=1
\end{equation}
Thus, the stream of attention and background can be denoted as :

\begin{equation}
\begin{split}
f_{attention}=(f_{full},f_{mask})\bigotimes att^{+}\\
f_{background}=(f_{full},f_{mask})\bigotimes att^{-}
\end{split}
\end{equation}
where $\bigotimes$ means the spatial weighting operation.

\subsection{Loss network with region-level task-guided loss}

With the attention maps described in last subsection, we further introduce the region-level triplet loss to enhance contrastive feature learning . After the attention operation, features from three main streams can be denoted as $f_{full}$, $f_{attetion}$ and $f_{background}$, $f_{full}$, $f_{attetion}$ and $f_{background}$ are used to calculate region-level task-guided loss for two task: keep the content and style transfer.

Our loss network can be divided into two parts: Feature reconstruction loss(a) and Style reconstruction loss(b), feature reconstruction loss is denoted as $\ell_{feat}$ which is the summary of $\ell_{gc}$ and $\ell_{lc}$, meanwhile, style reconstruction loss is denoted as $\ell_{style}$ which is the summary of $\ell_{gs}$ and $\ell_{ls}$.

\subsubsection{Feature reconstruction loss}

Traditional feature reconstruction loss which known as content loss only takes the input image $I_{RGB}$ as input and try to minimize the loss between the content of input image $I_{RGB}$ and output image $O_{RGB}$ without considering encoding content reconstructions. We address this problem with the image segmentation masks $I_{mask}$ for the input images, the local feature of pupil region can be addressed when calculating the loss of $f_{full}$ and $f_{attetion}$. To visualise the image information that is encoded at different layers of the input image with masks, we perform gradient descent on a white noise image to find another image that matches the feature responses of the original image with mask. We then define the squared-error loss between the two feature representations

 \begin{equation}
\ell_{feat}^l=\lambda_{g}\ell_{gc}^l+\lambda_{l}\ell_{lc}^l
\end{equation}

 \begin{equation}
\ell_{gc}^l=\sum_{c=1}^{C}\frac{1}{2N_{l}M_{l}}\sum_{ij}(F_{full}^{l,c}[O]-F_{full}^{l,c}[I])^{2}_{ij}
\end{equation}

 \begin{equation}
  \begin{split}
\ell_{lc}^l=\sum_{c=1}^{C}\frac{1}{2N_{l}M_{l}}\sum_{ij}(F_{full}^{l,c}[O]*F_{attention}^{l,c}[I]\\
-F_{full}^{l,c}[I]*F_{attention}^{l,c}[I])^{2}_{ij}
  \end{split}
\end{equation}
where $C$ is the number of channels in the semantic segmentation mask and $l$ indicates the $l$-th convolutional layer of the deep convolutional neural network, $F_{full}[\cdot]$ is the $f_{full}$ in each layer $l$ with the channel $c$, $F_{attention}[\cdot]$ is the $f_{attention}$ in each layer $l$ with the channel $c$, $\lambda_{g}$ is the weight to configure layer preferences of global losses, $\lambda_{l}$ is the weight to configure layer preferences of local losses.

Each layer with $N_{l}$ distinct filters has $N_{l}$ feature maps each of size $M_{l}$, where $M_{l}$ is the height times the width of the feature map. So the responses in each layer $l$ can be stored in a matrix $F[\cdot] \in R^{N_{l}\times M_{l}}$ where $F[\cdot]_{ij}$ is the activation of the $i^{th}$ filter at position $j$ in each layer $l$. As minimizing $\ell_{feat}$, the image content and overall spatial structure are preserved but color, texture, and exact shape are not.

\subsubsection{Style reconstruction loss}

Feature Gram matrices are effective at representing texture, because they capture global statistics across the image due to spatial averaging. Since textures are static, averaging over positions is required and makes Gram matrices fully blind to the global arrangement of objects inside the reference image. So if we want to keep the global arrangement of objects, make the gram matrices more controllable to compute over the exact region of entire image, we need to add some texture information to the image.

Instead of taking input image $I_{RGB}$ and style image $S_{RGB}$ as inputs, we take the input image $I_{RGB}$ and style image $S_{RGB}$ with their mask $I_{mask}$ and $S_{mask}$ as pair inputs. To learn the skin style and pupil style respectively, we denote the pupil region as attention region and extract attention maps $f_{attention}$ from both style image and input image, meanwhile, the skin region denoted as background region and product background maps $f_{background}$ from style image and input image. We then define the squared-error loss between the two region feature representations

\begin{equation}
\ell_{style}^l=\lambda_{g}\ell_{gs}^l+\lambda_{l}\ell_{ls}^l
\end{equation}
 \begin{equation}
\ell_{gs}^l=\sum_{c=1}^{C}\frac{1}{4N^2_{l,c}M^2_{l,c}}\sum_{ij}\left(G_{bkgd}^{l,c}[O]-G_{bkgd}^{l,c}[S]\right)^2_{ij}
\end{equation}
\begin{equation}
\ell_{ls}^l=\sum_{c=1}^{C}\frac{1}{4N^2_{l,c}M^2_{l,c}}\sum_{ij}\left(G_{attention}^{l,c}[O]-G_{attention}^{l,c}[S]\right)^2_{ij}
\end{equation}
where $C$ is the number of channels in the semantic segmentation mask and $l$ indicates the $l$-th convolutional layer of the deep convolutional neural network. Each layer with $N_{l}$ distinct filters has $N_{l}$ feature maps each of size $M_{l}$, where $M_{l}$ is the height times the width of the feature map. So the responses in each layer $l$ can be stored in a matrix $F[\cdot] \in R^{N_{l}\times M_{l}}$ where $F[\cdot]_{ij}$ is the activation of the $i^{th}$ filter at position $j$ in each layer $l$. $G^{l,c}[\cdot]$ be denoted as follows:

\begin{equation}
G_{bkgd}^{l,c}[O]=(F_{full}^{l,c}[O]*F_{bkgd}^{l,c}[I])*(F_{full}^{l,c}[O]*F_{bkgd}^{l,c}[I])^{T}
\end{equation}
\begin{equation}
G_{bkgd}^{l,c}[S]=(F_{full}^{l,c}[S]*F_{bkgd}^{l,c}[S])*(F_{full}^{l,c}[S]*F_{bkgd}^{l,c}[S])^{T}
\end{equation}
\begin{equation}
G_{attention}^{l,c}[O]=(F_{full}^{l,c}[O]*F_{attention}^{l,c}[I])*(F_{full}^{l,c}[O]*F_{attention}^{l,c}[I])^{T}
\end{equation}
\begin{equation}
G_{attention}^{l,c}[S]=(F_{full}^{l,c}[S]*F_{attention}^{l,c}[S])*(F_{full}^{l,c}[S]*F_{attention}^{l,c}[S])^{T}
\end{equation}
where$F_{full}[\cdot]$ is the $f_{full}$ in each layer $l$ with the channel $c$, $F_{attention}[\cdot]$ is the $f_{attention}$ in each layer $l$ with the channel $c$, $F_{background}[\cdot]$ is the $f_{background}$ in each layer $l$ with the channel $c$, $\lambda_{g}$ is the weight to configure layer preferences of global losses, $\lambda_{l}$ is the weight to configure layer preferences of local losses.

We formulate the style transfer objective by combining both two components together:
\begin{equation}
L_{total}=\sum_{l=1}^L\alpha_l\ell_{feat}^l+\sum_{l=1}^L\beta_l\ell_{style}^l
\end{equation}
where L is the total number of convolutional layers and $l$ indicates the $l$-th convolutional layer of the deep convolutional neural network. $\alpha_{l}$ and $\beta_{l}$ are the weights to configure layer preferences. $\ell_{feat}$ is the content loss (Eq.(10)) and $\ell_{style}$ is the style loss(Eq.(13)). $\alpha_l$,$\beta_l$ are scalars, $\alpha_l=10^{2}$,$\beta_l=10^{4}$, in all cases the hyperparameters $\alpha_l$,$\beta_l$ are exactly the same.  We find that unconstrained optimization of Equation 18 typically results in images whose pixels fall outside the range [0,255]. For a more fair comparison with our method whose output is constrained to this range, for the baseline we minimize Equation 18 using projected L-BFGS. Image O is generated by solving the problem
 \begin{equation}
O=\arg\min_{I}{L_{total}}+\theta\ell_{TV}(I)
\end{equation}
where I is initialized with white noise. The advantage of this solution is that the requirement for mask is not too precise. It does not only retain the desired structural features, but also enhance the estimation of the pupil and iris information during the reconstruction of the style.
  \begin{figure*}[!htbp]
\centering
 \begin{minipage}[]{1\textwidth}
    \centering
     \includegraphics[width = 1\textwidth,angle=0]{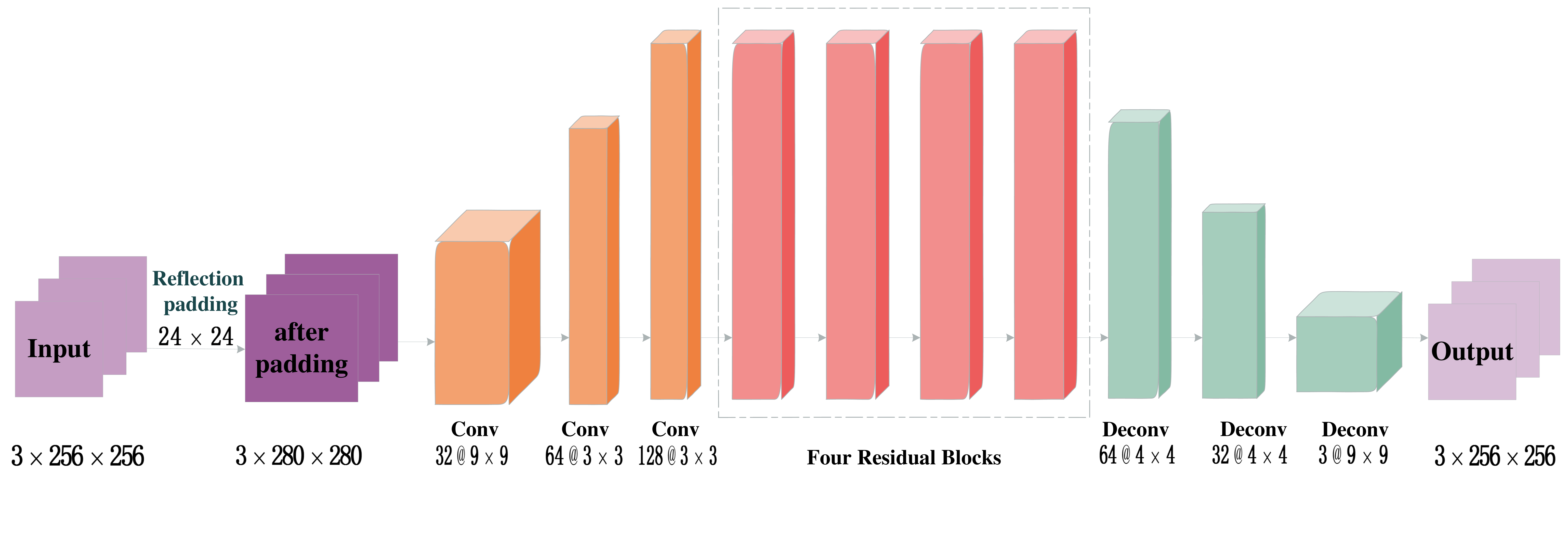}
 \end{minipage}
    \caption{The overview of feature extraction network. The first and last layers use $9\times9$ kernels, all other deconvolutional layers use $4\times4$ kernels with padding 1 and convolutional layers use $3\times3$ kernels with padding 0.}
    \label{figfeature}
\end{figure*}

\subsection{Feature extraction network}

\begin{figure*}[!htbp]
\centering
 \begin{minipage}[]{1\textwidth}
    \centering
     \includegraphics[width = 0.7\textwidth,angle=0]{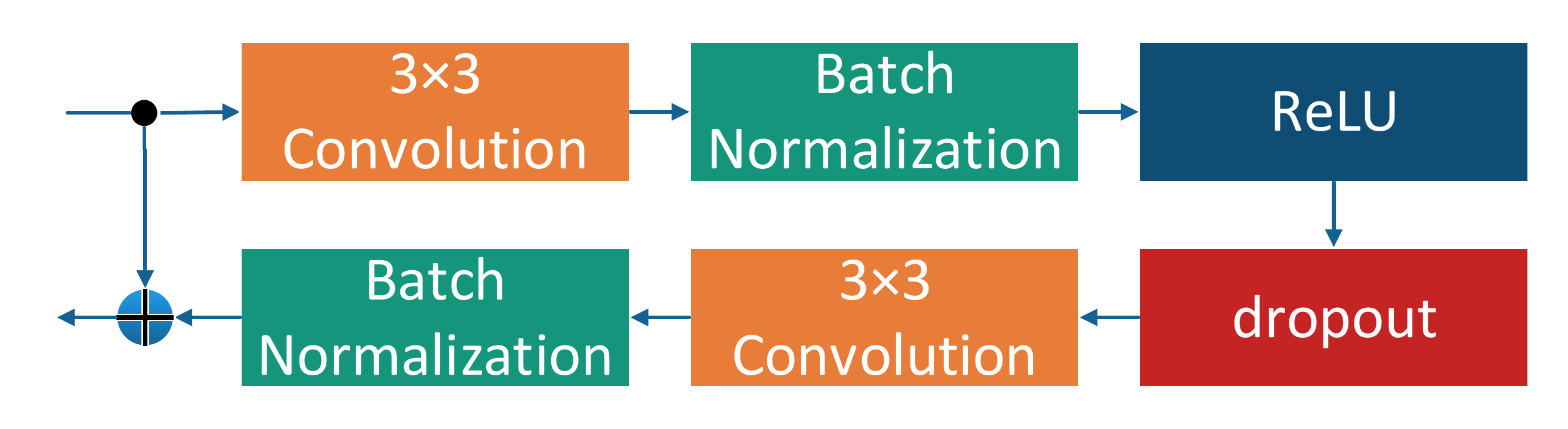}
 \end{minipage}
    \caption{The structure of residual block. Our residual blocks each contain two $3\times3$ convolutional layers with the same number of filters on both layer, similar with \cite{lifeifei2016} but with dropout followed by spatial batch normalization and a ReLU nonlinearity in order to avoid overfitting.}
    \label{figresidual}
 \end{figure*}

Our feature extraction network roughly follow the architectural guidelines set forth by \cite{DCGAN2015}. However, from \cite{checkboard2016} we know that the standard approach of producing images with deconvolution has some conceptually simple issues that lead to artifacts in produced images. Inspired by \cite{checkboard2016}, we modified the structure of image transformation network\cite{lifeifei2016} to our feature extraction network. The structure can be shown as Fig.\ref{figfeature}. The first and last layers use $9\times9$ kernels, all other deconvolutional layers use $4\times4$ kernels with padding 1 and convolutional layers use $3\times3$ kernels with padding 0. We use the residual block design similar with \cite{lifeifei2016} but with dropout followed by spatial batch normalization and a ReLU nonlinearity in order to avoid overfitting, shown in the Fig.\ref{figresidual}.Our network body comprises four residual blocks. All nonresidual convolutional layers are followed by batch normalization and ReLU nonlinearities with the exception of the output layer, which instead uses a scaled tanh to ensure that the output has pixels in the range [0,255].

\section{Experimental Results}\label{sec:experimentalresults}

We experimented with two tasks: style transfer and appearance-based gaze estimation. Previous style style transfer work has used optimization to generate images; our feed-forward structure gives similar qualitative results, but the speed is increased by three orders of magnitude. Previous work on appearance-based gaze estimation has used fine synthetic images for training and real images for testing, or training with real images and testing with fine synthetic images. By using simulated data or purified real data for training, and using purified real data for testing, we can get encouraging qualitative and quantitative results.

\subsection{Style Transfer}

The purpose of the style transfer is to generate an image that combines the content of the target content image as the real image content with the style of the target style image as the style of the synthetic image. We train an image transformation network for each of the several hand selection style goals and compare our results with the baseline methods of Gatys et al.\cite{Gatys2015} and Feifei Li et al.\cite{lifeifei2016}. As a baseline, we re-implemented the method of Gatys et al.\cite{Gatys2015} and Feifei Li et al.\cite{lifeifei2016}. In order to make a fairer comparison with our method whose output is constrained to [0, 255], for the baseline, we minimize the equation 3 and equation 4 by using the projected L-BFGS by cropping the image to the range [0, 255] at each iteration. In most cases, the optimization converges to satisfactory results in 500 iterations.

$\mathbf{Implementation}$ $\mathbf{Details}$: We resize each of the 80 thousand training images to $256 \times 256$ and train our network with a batch size of 4 for 50000 iterations, giving roughly two epochs over the training data. We use Adam with a learning rate of $1 \times 10^{-4}$. The output images are regularized with total variation regularization with a strength of between $1 \times 10^{-7}$ and $1 \times 10^{-5}$. We choose conv$4\_2$  as the local content representation, and conv$1\_1$, conv$2\_1$, conv$3\_1$, conv$4\_1$ and conv$5\_1$ as the local style representation. conv$3\_2$  as the global content representation, and conv$1\_2$, conv$2\_2$, conv$3\_3$, conv$4\_3$ and conv$5\_3$ as the global style representation. Our implementation use Torch7 and cuDNN, training takes roughly 3 hours on a single GTX Titan X GPU.

\begin{figure*}[!htbp]
\centering
 \begin{minipage}[]{1\textwidth}
    \centering
     \includegraphics[width = 1\textwidth,angle=0]{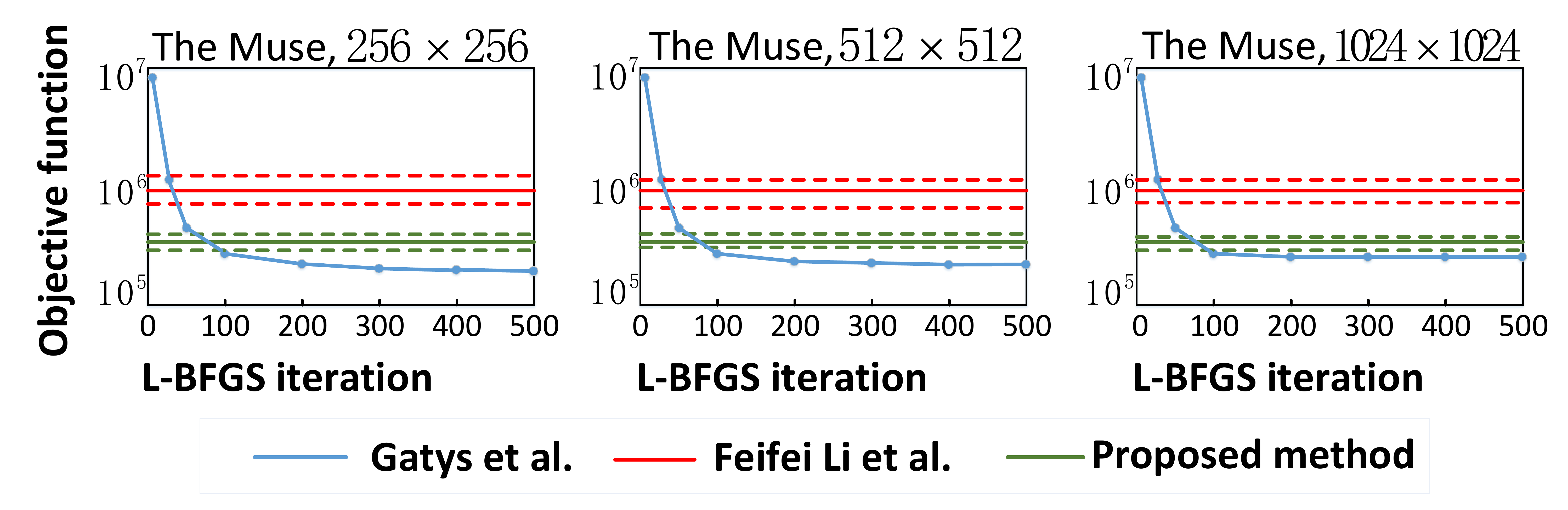}
 \end{minipage}
    \caption{Proposed style transfer networks and Gatys et al.\cite{Gatys2015} and Feifei Li et al.\cite{lifeifei2016} minimize the same objects. We compare their object values on 50 images; dashed lines and error bars show standard deviations. Our networks are trained on $256 \times 256$ images but generalize to larger images. }
    \label{fig10}
\end{figure*}

$\mathbf{Qualitative}$ $\mathbf{Results}$: Fig.7 describes the style transfer method proposed in comparison to methods proposed by  Gatys et al.\cite{Gatys2015} and Feifei Li et al.\cite{lifeifei2016} across series of indoor and outdoor scenes from UnityEyes\cite{Wood2016learning} and LPW\cite{LPW2016} datasets respectively. (a),(b),(c),(d),(e), and (f) represents six different conditions of outdoor scenes from LPW dataset. On the other hand, styles A, B, and C from the UnityEyes dataset represent three different distributions of indoor conditions, which if closely observed, it can be seen that none of these styles has similar gaze angle with real images.

From (a),(b), and (c), it can be observed that the proposed method is less affected by light and achieves similar results with Gatys et al. \cite{Gatys2015} and Feifei Li et al.\cite{lifeifei2016}, but the proposed method can better preserve the color information of style image. From (d),(e), and (f), it can be seen that Gatys et al.\cite{Gatys2015} and Feifei Li et al.\cite{lifeifei2016} are influenced by light and other factors, the pupil and the iris cannot be completely separated. What's more, the distribution of pupil and iris regions is dramatically different from style image. The proposed method, therefore can separate the pupil and the iris regions easily and the distribution of pupil and iris regions is similar to style image.

Furthermore, it can be observed that no matter how style image changes, the distribution of the purified image is more inclined to that of the style image, which changes slightly according to different style images. However, the distribution of Gatys et al.\cite{Gatys2015}, Feifei Li et al.\cite{lifeifei2016} is more complex, because of light and other external factor interference, making it difficult to learn for gaze estimation tasks. Note that the proposed method preserves the annotation information while purifying the illumination of the real images.

\begin{figure*}[!htbp]
\centering
 \begin{minipage}[]{1\textwidth}
    \centering
     \includegraphics[width = 1\textwidth,angle=0]{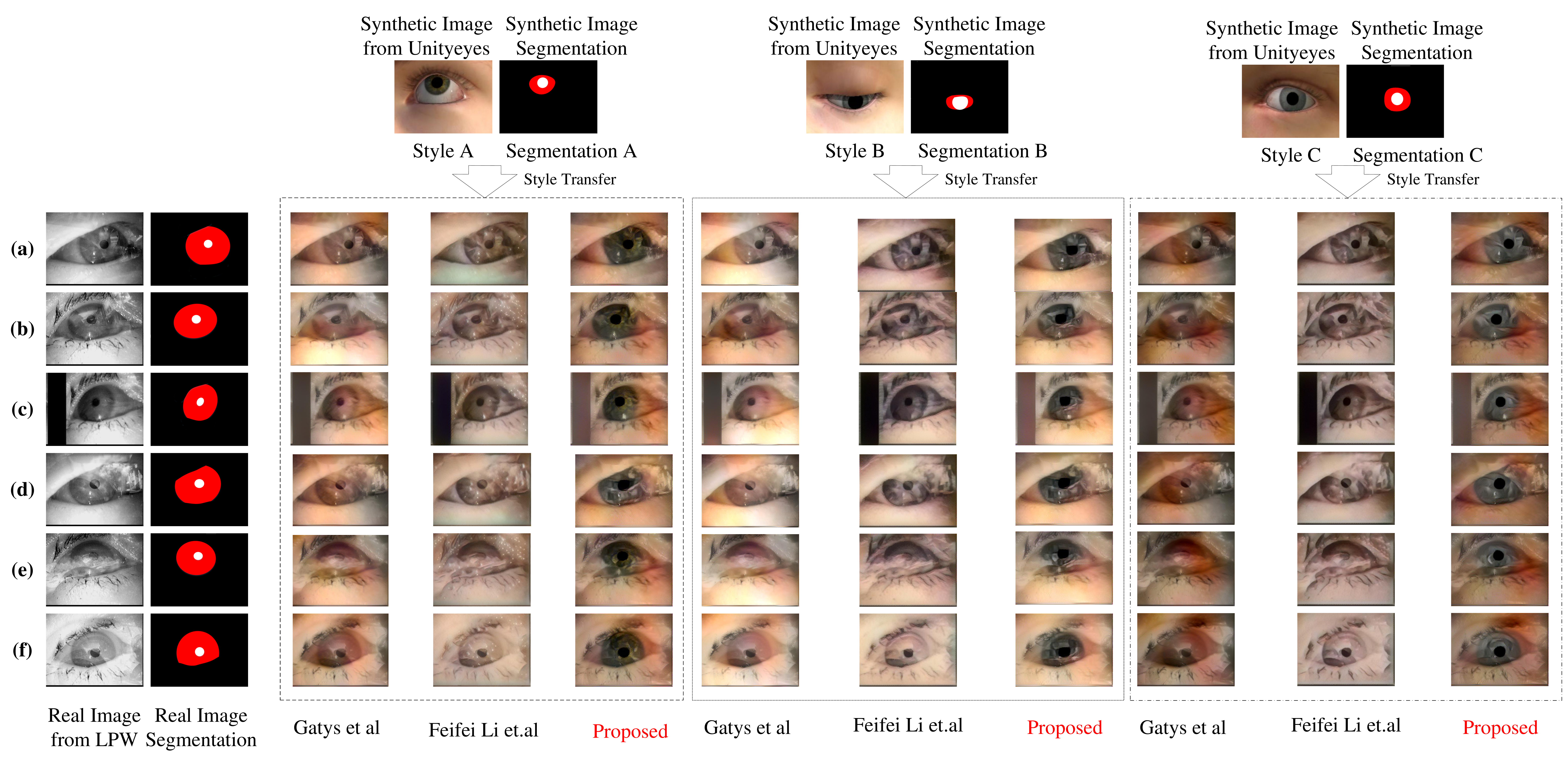}
 \end{minipage}
    \caption{Comparison on public LPW dataset with available style transfer methods.(a),(b),(c),(d),(e),and (f) represent the purified results of different distributions under six outdoor conditions from LPW dataset with three different styles from UnityEyes dataset. Style A, B, and C represent three different distributions of indoor conditions. The distribution of pupil and iris regions is dramatically different from style image. The proposed method, therefore can separate the pupil and the iris regions easily and the distribution of pupil and iris regions is similar to style image. }
    \label{fig1_1}
\end{figure*}

$\mathbf{Quantitative }$ $\mathbf{Results}$: As evidenced by Gatys et al.\cite{Gatys2015} and Feifei Li et al.\cite{lifeifei2016} and reproduced in Figure 8, the image that produces the minimized pattern reconstruction loss preserves the style characteristics of the target image, but does not preserve its spatial structure. Reconstruction from higher layers transfers large-scale structures from the target image.
The baseline and our methods both minimize equation 19. The baseline performs explicit optimization over the output image, while our method is trained to find a solution for any content image in a single forward pass. We may therefore quantitatively compare the two methods by measuring the degree to which they successfully minimize Equation 19.

We used Pablo Picasso's Muse as a style image to run our method and baseline method on 50 images of the MS-COCO validation set. For the baseline method, we record the value of the objective function for each optimization iteration. For our method, we record Equation 19 for each image. From Figure 9, we can see that Feifei Li et al.\cite{lifeifei2016} achieved high losses, and our method achieved a loss comparable to 0 to 80 explicit optimization iterations.

Although our networks are trained to minimize Equation 19 for $256\times256$ images, they  are also successful in minimizing the objective when applied to larger images. We repeat the same quantitative evaluation for 50 images at $512\times512$ and $1024\times1024$, results are shown in figure 9. We can see that even at higher resolutions our method achieves a loss comparable to 50 to 100 iterations of the baseline method.

\begin{table}[!hbp]
\centering
\label{tabLr}
\caption{ Speed (in seconds) for our style transfer networks vs Gatys et al. \cite{Gatys2015}, Feifei Li et al.\cite{lifeifei2016} for various resolutions. Across all image sizes, compared to 400 iterations of the baseline method, our method is three orders of magnitude faster than Gatys et al. \cite{Gatys2015} and we achieve better qualitative results (Fig.\ref{fig1_1} compared with Feifei Li et al.\cite{lifeifei2016} in tolerate speed. Our method processes $512\times512$ images at 20 FPS, making it feasible to run in real-time or on video. All benchmarks use a Titan X GPU.}
\begin{tabular}{|c|c|c|c|}
\hline
\diagbox{Method}{Time}{Resolution} & $256\times256$ & $512\times512$ & $1024\times1024$\\
\hline
Gatys et al. &12.69s & 45.88s & 171.55s\\
\hline
 Feifei li et al. & 0.023s & 0.08s & 0.35s \\
\hline
$\mathbf{Proposed Method}$ & $\mathbf{0.015s}$ &$\mathbf{0.05s}$ & $\mathbf{0.21s}$ \\
\hline
speedup (proposed vs Gatys) & $\mathbf{1060x}$ & $\mathbf{1026x}$ & $\mathbf{1042x}$ \\
\hline
speedup (proposed vs Feifei Li) & $\mathbf{1.53x}$ & $\mathbf{1.6x}$ & $\mathbf{1.67x}$ \\
\hline
\end{tabular}
\end{table}

$\mathbf{Speed}$: Table 2 compares the runtime of our method and Gatys et al.\cite{Gatys2015}, Feifei Li et al.\cite{lifeifei2016} for several image sizes. Across all image sizes, compared to 400 iterations of the baseline method, our method is three orders of magnitude faster than Gatys et al.\cite{Gatys2015} and we achieve better qualitative results (Fig.\ref{fig1_1}) compared with Feifei Li et al.\cite{lifeifei2016} in tolerate speed. Our method processes images of size $512\times512$  at 20 FPS, making it feasible to run in real-time or on video.

\subsection{Appearance-based Gaze Estimation}

We evaluate our method for appearance-based gaze estimation on the MPIIGaze and purified MPIIGaze with base-line methods.
\begin{figure*}[!htbp]
\centering
 \begin{minipage}[]{1\textwidth}
    \centering
     \includegraphics[width = 1\textwidth,angle=0]{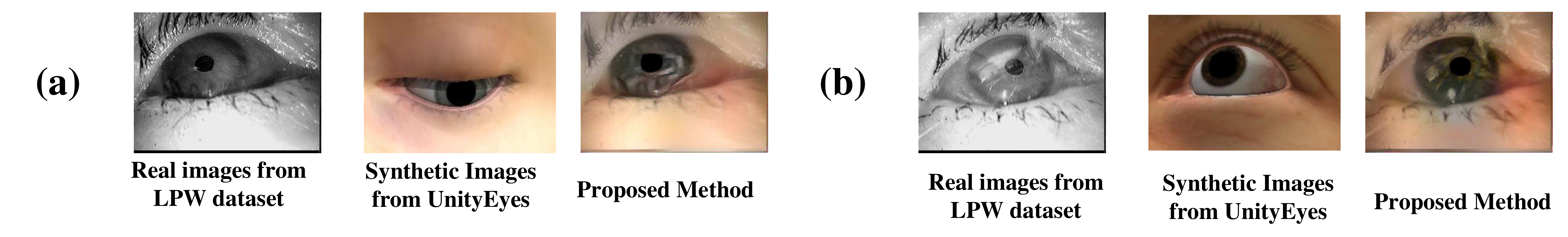}
 \end{minipage}
    \caption{Example output of proposed method for the LPW gaze estimation dataset. The skin texture and the iris region in the purified real images are qualitatively significantly more similar to the synthetic images than to the real images. }
    \label{fig11}
\end{figure*}

$\mathbf{Implementation }$ $\mathbf{Details}$: In order to verify the effectiveness of the proposed method for gaze estimation, 3 public datasets (UTView\cite{Sugano2014learning}, SynthesEyes\cite{Wood2015rendering}, UnityEyes\cite{Wood2016learning}) are used to train the estimator with k-NN\cite{wang2017}, MPIIGaze dataset\cite{MPIIGaze2017} and purified MPIIGaze dataset (purified by proposed method) are used for test the accuracy. The eye gaze estimation network is similar to \cite{IJCNN2018}\cite{PR2018}, the input is a $35 \times 55$ gray scale image that is passed through 5 convolutional layers followed by 3 fully connected layers, the last one encoding the 3-dimensional gaze vector: (1)Conv 32$@$3$\times$3 (2)Conv 32$@$3$\times$3 (3)Conv 64$@$3$\times$3 (4)Max-Pooling 3$\times$3 (5)Conv 80$@$3$\times$3 (6)Conv 192$@$3$\times$3 (7)Max-Pooling 2$\times$2 (8)FC9600 (9)FC1000 (10)FC3 (11)Euclidean loss. All networks are trained with a constant $1\time10^{-3}$ learning rate and 512 batch size, until the validation error converges.

$\mathbf{Qualitative }$ $\mathbf{Results}$: Fig.9 shows examples of real, synthetic and purified real images from the eye gaze dataset. As shown, we observe a significant qualitative improvement of real images: Proposed method successfully captures the skin texture, sensor noise and the appearance of the iris region in the synthetic images. Note that our method preserves the annotation information(gaze direction) while purifying the illumination.

\begin{table}[!hbp]
\centering
\label{tabLr1}
\caption{Test performance on MPIIGaze and purified MPIIGaze; Purified MPIIGaze is the dataset which purified by proposed method. "Method" represents training set used with gaze estimation method.  Note how purifying real dataset for training lead to improved performance. }
\begin{tabular}{|c|c|c|}
\hline
Method & MPIIGaze & purified MPIIGaze \\
\hline
Support Vector Regression(SVR) &16.5$^{\circ}$ & 14.3$^{\circ}$ \\
\hline
Adaptive Linear Regression(ALR) &16.4$^{\circ}$ & 13.9$^{\circ}$ \\
\hline
Random Forest(RF) & 15.4$^{\circ}$ & 14.2$^{\circ}$ \\
\hline
KNN with UTview &16.2$^{\circ}$ & 13.6$^{\circ}$ \\
\hline
CNN with UTview &13.9$^{\circ}$ & 11.7$^{\circ}$ \\
\hline
KNN with UnityEyes  &12.5$^{\circ}$ & 9.9$^{\circ}$ \\
\hline
CNN with UnityEyes &9.9$^{\circ}$ & 7.8$^{\circ}$ \\
\hline
KNN with Syntheyes  &11.4$^{\circ}$ &8.0$^{\circ}$  \\
\hline
CNN with Syntheyes  &13.5$^{\circ}$ & 8.8$^{\circ}$ \\
\hline
\end{tabular}
\end{table}
\begin{figure*}[!htbp]
\centering
 \begin{minipage}[]{1\textwidth}
    \centering
     \includegraphics[width = 0.8\textwidth,angle=0]{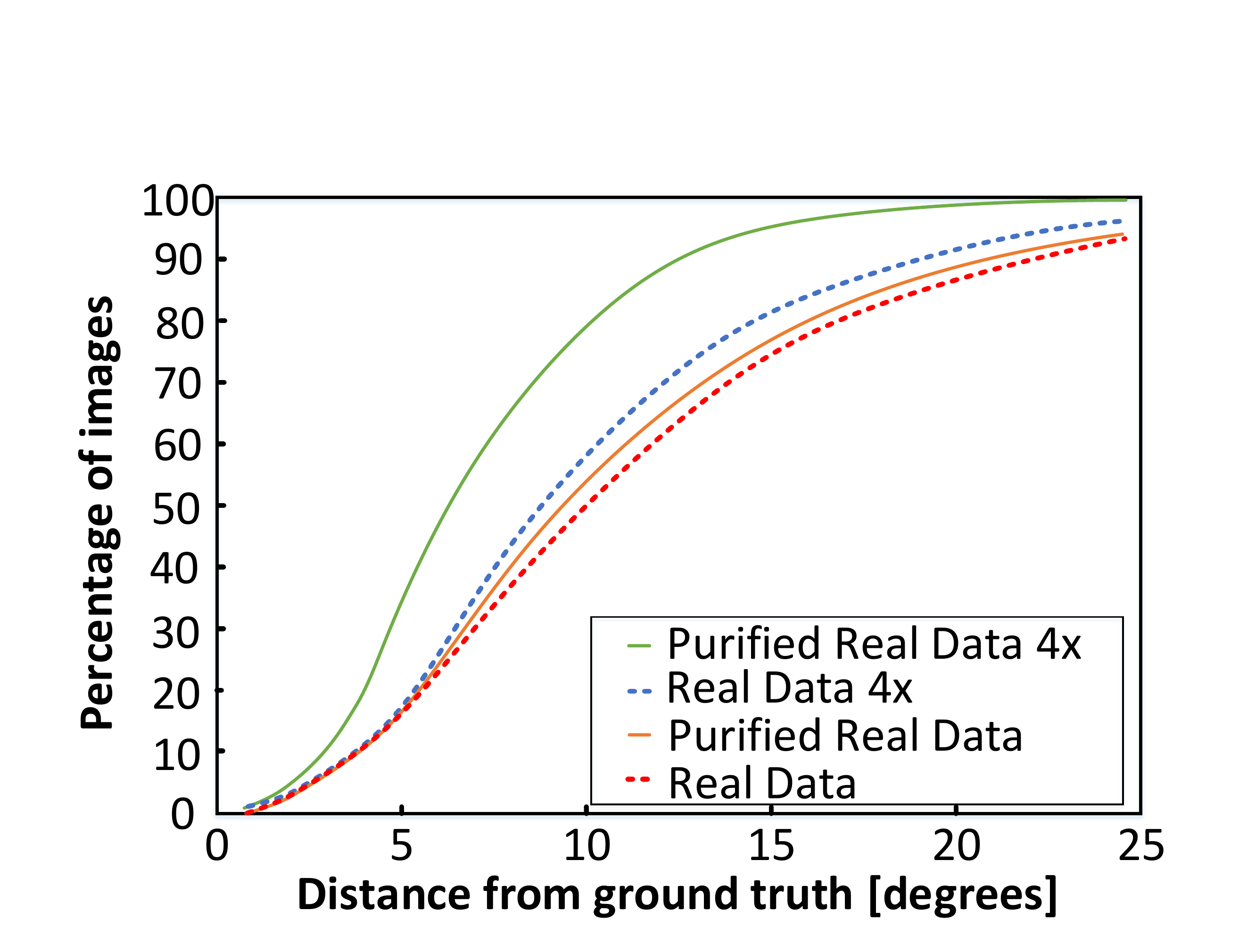}
 \end{minipage}
    \caption{Quantitative results for appearance-based gaze estimation on the MPIIGaze dataset and purified MPIIGaze dataset. The plot shows cumulative curves as a function of degree error as compare to the ground truth eye gaze direction, for different numbers of testing examples of data.}
    \label{fig12}
\end{figure*}
$\mathbf{Quantitative }$ $\mathbf{Results}$: Five gaze estimation methods are used as base-line estimation methods. In addition to common methods such as Support Vector Regression(SVR), Adaptive Linear Regression(ALR) and Random Forest(RF), two methods are reproduced for fairly comparison with state-of-the-art. First method is a simple cascaded method\cite{wang2017}\cite{ACML2018}\cite{ICIMCS2017} which uses multiple $k$-NN($k$-Nearest Neighbor) classifier to select neighbors in feature space joint head pose,pupil center and eye appearance. The other method is to train a simple convolutional neural network (CNN)\cite{wild2017}\cite{IJCNN2018}\cite{PR2018} to predict the eye gaze direction with $ l_{2}$ loss. We train on UnityEyes ,UTView, SynthesEyes and test on MPIIGaze, purified MPIIGaze which is purified by proposed method. When testing on the MPIIGaze dataset, the training data can be either a raw dataset or a synthetic dataset, and when tested on a purified MPIIGaze dataset, the training data is either a purified real dataset or a raw synthetic dataset. Table 3 compares the performance of these two gaze estimation methods with different datasets. "Method" represents training set used with gaze estimation method. Large improvement in performance of testing on the output of proposed method is observed, each dataset improves at least three degrees of gaze estimation accuracy. This improvement shows the practical value of our method in many HCI tasks.

$\mathbf{Preserving }$ $\mathbf{Ground}$ $\mathbf{Truth}$: To quantify that the ground truth gaze direction doesn't change significantly, we manually labeled the ground truth pupil centers in 200 real and purified images by fitting an ellipse to the pupil. This is an approximation of the gaze direction, which is difficult for humans to label accurately. The absolute difference between the estimated pupil center of real and corresponding purified images is quite small: 0.8 $\pm$ 1.1 (eye width=55px)

\section{Conclusion}\label{sec:conclusion}

This paper took the first step to purify the real image by weakening its distribution, which is a better choice than improving the realism of synthetic image. We have applied this method to style transfer and gaze estimation tasks where we achieved comparable performance and drastically improved speed compared to existing methods. Performance evaluation indicates that purified MPIIGaze dataset (purified by our proposed method) recorded smaller error angle when used for gaze estimation task as compared with the raw MPIIGaze dataset.

In future, we intend to explore modeling the real-time gaze estimation system based on the proposed method and improve the speed of purifying videos.





\section*{ACKNOWLEDGMENTS}

The authors sincerely thank the editors and anonymous reviewers for the very helpful and kind comments to assist in improving the presentation of our paper. This work was supported in part by the National Natural   Science Foundation of China Grant 61370142 and Grant 61802043, by the   Fundamental Research Funds for the Central Universities Grant 3132016352,   by the Fundamental Research of Ministry of Transport of P. R. China Grant   2015329225300.

\end{document}